\documentclass[11pt]{article}

\usepackage[final]{acl}
\usepackage{hyperref}
\usepackage{fontawesome5} 
\usepackage{times}
\usepackage{latexsym}
\usepackage{amsfonts}
\usepackage{enumitem}
 \usepackage{amsmath} 
\usepackage[T1]{fontenc}

\usepackage[utf8]{inputenc}

\usepackage{microtype}

\usepackage{inconsolata}

\usepackage{graphicx}

\title{The Proxy Presumption: From Semantic Embeddings to Valid Social Measures}

\author{
 \textbf{Baishi Li\textsuperscript{1}},
 \textbf{Ta Yu\textsuperscript{1}},
 \textbf{Kelvin J.L. Koa\textsuperscript{1,2}}\thanks{Corresponding author.},
 \textbf{Ke-Wei Huang\textsuperscript{1,2}}
\\
\\
 \textsuperscript{1}National University of Singapore,
 \textsuperscript{2}Asian Institute of Digital Finance
\\
{\small
\faEnvelope\,: 
\href{mailto:baishili@u.nus.edu}{\texttt{baishili@u.nus.edu}},\quad 
\href{mailto:ta.yu@u.nus.edu}{\texttt{ta.yu@u.nus.edu}},\quad 
\href{mailto:kelvin.koa@u.nus.edu}{\texttt{kelvin.koa@u.nus.edu}},\quad 
\href{mailto:dishkw@nus.edu.sg}{\texttt{dishkw@nus.edu.sg}}
}
}

\begin{document}
\maketitle
\begin{abstract}
Natural Language Processing is rapidly evolving into a primary instrument for Computational Social Science, with researchers increasingly using embeddings to measure latent constructs such as \textit{novelty}, \textit{creativity}, and \textit{bias}. However, this transition faces a fundamental validity challenge: the ``Proxy Presumption,'' or the reliance on geometric properties (e.g., cosine distance) as direct measures of social concepts. We argue that without explicit validation, unsupervised representations remain entangled mixtures of the target construct ($C$) and confounding attributes ($Z$) like topic, style, and authorship. To bridge the gap between semantic embeddings and valid social measures, we introduce the Construct Validity Protocol (CVP). Drawing on causal representation learning and psychometrics, the CVP offers a rigorous pipeline from conceptualization to quantitative verification. We further propose Counterfactual Neutralization, a novel method using LLMs to reduce confounding in embedding space. By providing a standardized Validity Suite---including tests for discriminant, incremental, and predictive validity---this work offers the community a toolkit to transform heuristic proxies into robust, scientifically defensible instruments.
\end{abstract}

\section{Introduction}
The core ambition of contemporary Natural Language Processing (NLP) has expanded beyond predicting the next token; we are increasingly tasked with measuring the social world. Recent literature has pioneered the quantification of abstract social constructs, proposing computational metrics for concepts such as \textit{novelty}, \textit{creativity}, and \textit{bias} \citep{merrill-etal-2024-evaluating,lee-etal-2024-icscore,bang-etal-2024-measuring}. This transition marks a pivotal moment: NLP is evolving from an engineering discipline into a primary measurement instrument for Computational Social Science (CSS).

However, this ambition introduces a methodological challenge: the gap between \textbf{theoretical constructs} (what we intend to measure) and \textbf{computational proxies} (what we implement). In the absence of established ground truth for latent social variables, the field faces a risk we term the \textit{Proxy Presumption}: a convenient geometric property---most commonly cosine similarity in embedding space---is renamed as a theoretical construct without sufficient validation \citep{caliskan2017semantics,may2019seat,bolukbasi2016debias,garg2018stereotypes,kozlowski2019geometry}. We emphasize that this is a recurring pattern across a substantial and influential subset of NLP/ML measurement work rather than an isolated mistake, while recognizing meaningful variation in how rigorously different papers validate their proxies. While vector distance captures semantic divergence, equating it directly with ``innovation'' or ``creativity'' assumes a strong isomorphism between embedding space geometry and social reality that may not hold.

While measurement validity is a general issue in machine learning, the proxy presumption is especially acute in NLP because text is the primary unstructured medium for computational social science, and major nuisance dimensions ($Z$) in text are inherently linguistic (register, dialect, pragmatics, style). Moreover, many interventions that can reduce confounding are text-native, including counterfactual rewriting and language-aware debiasing methods \citep{elazar2018adversarial,ravfogel2020,blodgett2020,sap2019risk}.

This paper argues that such ``measurement by renaming'' can be formalized as a non-identification problem: unsupervised embeddings compress multiple generative factors into a single representation, entangling the target construct ($C$) with nuisance attributes ($Z$) such as topic, author style, register, or length \cite{locatello2019,scholkopf2021}. Without explicit disentanglement or design-based controls, standard metrics may primarily capture method variance rather than the intended social phenomenon \cite{zhou2022}. For example, a high cosine distance may reflect a genuine conceptual break, or it may simply reflect a vocabulary shift. Even with perfect disentanglement, the current practice still suffers from the mis-specification issue: the function (such as cosine similarity) chosen to represent the new concept may not be the most suitable function that represents the functional relationship between $C$ and the document, excluding the influences from confounding covariates $Z$.

To address this challenge and support the maturation of CSS, we introduce the \textbf{Construct Validity Protocol (CVP)}. Adapted from standards in psychometrics and causal representation learning \cite{bengio2013,locatello2019,scholkopf2021}, the CVP provides a roadmap for validating social variables in NLP. Our contributions are fourfold:
\begin{enumerate}
    \item \textbf{A non-identification result:} We formalize why unsupervised metrics cannot reliably recover latent constructs without explicit assumptions or interventions \cite{locatello2019,scholkopf2021}.
    \item \textbf{Methodological solutions:} We propose \textbf{Counterfactual Neutralization} and connect it to existing debiasing/disentanglement tools such as adversarial removal and nullspace projection \cite{elazar2018adversarial,ravfogel2020}.
    \item \textbf{A Validity Suite:} We define a standardized suite of evaluations (stability/reliability, convergent validity, discriminant and incremental validity, known-groups checks, and criterion-related evidence) to test whether a proxy tracks the intended construct beyond nuisance dimensions.
    \item \textbf{A forensic literature review:} We analyze recent publications (2021--2025) to document the reuse of nearly identical similarity-based instruments for distinct, sometimes contradictory, constructs.
\end{enumerate}

\paragraph{Scope and positioning.}
This paper is a \emph{position and synthesis} contribution: we propose CVP as a community-facing reporting protocol for measurement identification, not a final or exclusive standard. We present CVP as a starting point that should be iterated as evidence accumulates and as NLP-specific failure modes are better understood, analogous in spirit to evolving reporting guidelines in adjacent areas (e.g., Data Statements for NLP; \citet{bender-friedman-2018-datastatements}).

Our goal is not to discourage the measurement of complex social phenomena, but to provide tools that render such measurements scientifically defensible. In Computational Social Science (CSS), these constructed variables are often critical inputs for downstream causal inference or predictive tasks. However, if the variable construction itself lacks validity, no causal ML methods can rigorously establish downstream causality.
\section{Related Literature: The State of Social Measurement}

Our work connects three research traditions that are often cited separately but rarely integrated in NLP: (i) measurement theory and construct validity from the social sciences, (ii) causal representation learning and identifiability, and (iii) recent NLP methodology debates about what models and annotations actually measure.

\subsection{Measurement Theory and Construct Validity.}
Measurement theory distinguishes \textit{constructs} (theoretical concepts) from \textit{measures} (observable indicators). As codified in standard social science guidelines \cite{devellis2016scale,adcock2001measurement}, validation is not a definitional act but an empirical one: researchers must demonstrate that a measure behaves as predicted within a ``nomological network,'' correlating with related concepts (convergent validity) while remaining distinct from nuisances (discriminant validity). This rigorous framework is increasingly relevant to NLP as the field targets complex social variables. A growing body of work now operationalizes constructs through explicit \textit{domains of observables}---for example, decomposing persuasion into specific resistance strategies or mapping social norms to defined interactional patterns \cite{socialgaze2024,cali2023}. These studies exemplify the ``construct-to-measure'' pipeline we formalize, standing in contrast to scalar metrics that prioritize prediction without separating signal from confounding noise.

\subsection{Validity in NLP: Annotation, Bias, and the Limits of Proxying.}
A parallel line of work argues that many NLP variables inherit ambiguity from their measurement process. Human label variation is increasingly recognized not as annotator error, but as a meaningful signal of linguistic ambiguity and subjectivity \citep{pavlick-kwiatkowski-2019-disagreements,plank-2022-labelvariation,davani2022disagreements,aroyo2015crowdtruth}. Furthermore, dataset curation and annotation protocols can embed structural artifacts and degenerate shortcuts into text, yielding variables that appear to measure a target construct while in fact tracking annotation or sampling regularities \citep{gururangan2018annotation,poliak2018hypothesis,mccoy2019right,bender-friedman-2018-datastatements,geiger2020garbage}. Methodological critiques in fairness and social NLP emphasize that quantities like ``toxicity'' and ``bias'' are often operationalized as convenient proxies that suffer from construct slippage and normative mismatch, including dialect and genre confounds \citep{blodgett2020,sap2019risk}. This perspective complements our claim that measurement failures are frequently \textit{method-variance failures}: what appears to be a social construct may largely reflect dataset composition, genre, dialect, or frequency artifacts. Our CVP makes these critiques actionable by requiring explicit confound tests, invariance checks aligned with established NLP evaluation practice \citep{ribeiro2020checklist,gardner2020contrast}, and transparent reporting of what the proxy is sensitive to.

\subsection{Causal Representation Learning and Non-Identification.}
We ground our framework in Causal Representation Learning \cite{scholkopf2021}, which establishes a fundamental non-identification result: without explicit structural assumptions or interventions, it is theoretically impossible to disentangle latent causal factors from observational data alone. We apply this constraint to social measurement: extracting a valid construct ($C$) from text requires actively modeling and neutralizing nuisance factors ($Z$), rather than assuming that an unsupervised embedding will spontaneously isolate them.

Beyond representation learning, a complementary literature studies when text-derived variables can support downstream causal inference \citep{egami2022causaltexts} and how to use imperfect surrogates for downstream inference via design-based estimators \citep{egami2023imperfectsurrogates}. Learned-proxy methodology in political science further emphasizes that post-hoc correlation is insufficient for testing causal theories with learned proxies \citep{knox2022learnedproxies}. We position CVP as addressing the measurement-validity prerequisite that these downstream frameworks typically assume.
\section{The Impossibility of Unsupervised Inverse Construction}

Current computational social science often relies on the assumption that a specific latent social construct $C$ (e.g., a hate sentiment score) can be recovered directly from observed documents $D$ using a fixed proxy function $f(D)$. We argue that this formulation constitutes an \textit{ill-posed inverse problem} formally analogous to the identification problem in causal inference.

\subsection{Case I: Single-Document Measurement}

Let the generation of a document $D \in \mathcal{D}$ be modeled as a stochastic process dependent on two disjoint sets of latent factors: a \textbf{scalar} target construct $c \in \mathbb{R}$ and a vector of nuisance attributes $\mathbf{z} \in \mathbb{R}^k$ (e.g., \textit{Topic}, \textit{Length}, \textit{Authorship}). We model the document generation process as a draw from a conditional probability distribution parameterized by $\theta$.
\begin{equation}
   p_\theta(D \mid c, \mathbf{z})
\end{equation}
The standard measurement pipeline attempts to invert this process:
\begin{enumerate}
    \item \textbf{Representation Learning:} An encoder $E: \mathcal{D} \to \mathbb{R}^d$ maps the text to an embedding space $\mathbf{e}$.
    \item \textbf{Proxy Construction:} A function $f: \mathbb{R}^d \to \mathbb{R}$ is applied to estimate the scalar construct: $\hat{c} = f(\mathbf{e})$.
\end{enumerate}

Ideally, $f(E(D))$ should recover $c$. However, as established in the representation learning literature \cite{locatello2019, scholkopf2021}, this recovery is theoretically impossible without structural assumptions.

\paragraph{Proposition 1 (Rotational Ambiguity).}
\textit{Let the joint latent space be $\mathbf{h} = [c; \mathbf{z}]$ with a factorized isotropic Gaussian prior $p(\mathbf{h}) = \mathcal{N}(0, I)$. For any unsupervised objective that maximizes the likelihood of the data $p(D)$, the learned representation is identifiable only up to an arbitrary orthogonal rotation. Consequently, the scalar dimension $c$ is inextricably mixed with the nuisance vector $\mathbf{z}$.}

\paragraph{Proof Sketch.}
Following \cite{locatello2019}, consider the marginal likelihood $p(D) = \int p_\theta(D \mid \mathbf{h}) p(\mathbf{h}) d\mathbf{h}$. Let $R$ be any arbitrary orthogonal matrix ($R^\top R = I$). We define a transformed latent space $\mathbf{h}' = R\mathbf{h}$. Due to the isotropy of the Gaussian prior, the density is rotation-invariant: $p(\mathbf{h}') = p(\mathbf{h})$.

Now, consider an alternative generator defined by $G'(\mathbf{x}) = G(R^\top \mathbf{x})$. If we assume the data generation process uses latent variables $\mathbf{h}'$ and generator $G'$, the observation is $D = G'(\mathbf{h}') = G(R^\top (R\mathbf{h})) = G(\mathbf{h})$. Thus, the model $(G', \mathbf{h}')$ yields the exact same observational distribution $p(D)$ as the original model $(G, \mathbf{h})$.

Since the unsupervised objective depends only on $p(D)$, it cannot distinguish between the entangled basis $\mathbf{h}'$ (where the first dimension is a linear combination of $c$ and $\mathbf{z}$) and the disentangled basis $\mathbf{h}$. In other words, any $\mathbf{h}' = \alpha c + \beta^\top \mathbf{z}$ is a possible latent factor vector. Thus, $c$ is not identified.

\paragraph{Implication.}
This proves that statistical independence does not imply disentanglement \cite{scholkopf2021}. Even if $c \perp \mathbf{z}$ in the real world, the learned embedding $\mathbf{e}$ can arbitrarily rotate the basis such that the dimension used for measurement is a mixture of signal and noise. Thus, any proxy $\hat{c} = f(\mathbf{e})$ suffers from intrinsic method variance. We will illustrate this claim with an example in the two-document case.

\subsection{Case II: Two-Document Measurement}

Constructing variables from pairwise document embeddings introduces an additional layer of entanglement. Consider measuring the relationship between two documents, such as \textit{Scientific Novelty} ($C$), defined as the distance between a new paper $D_i$ and prior work $D_j$. The key nuisance is topical overlap: both documents may share a dominant topic (e.g., ``Deep Learning''), which can account for most surface-level lexical and semantic mass. Standard practice then applies a similarity or distance function (e.g., cosine similarity) to the \emph{full} embeddings, implicitly allowing this nuisance component to dominate the measurement.

Even if we assume an oracle encoder $E$ that produces a perfectly disentangled concatenation $\mathbf{e}=[\mathbf{c}\oplus\mathbf{z}]$, where $\mathbf{c}$ encodes the concept of interest and $\mathbf{z}$ encodes nuisance variation, similarity-based measurement can still fail for two reasons:
\begin{enumerate}[label=\textbf{(\Roman*)}, leftmargin=*]
    \item \textbf{Target mismatch.} The metric of interest should be applied to the concept embeddings $(\mathbf{c}_i,\mathbf{c}_j)$, but is typically applied to the full vectors $(\mathbf{e}_i,\mathbf{e}_j)$, allowing variance in $\mathbf{z}$ to dominate.
    \item \textbf{Metric indeterminacy.} Even if $\mathbf{c}$ were perfectly isolated, there is no theoretical guarantee that a particular geometric function (e.g., cosine similarity versus Euclidean distance) is the correct proxy for an abstract construct. Determining the appropriate functional form requires empirical validation (Section~4).
\end{enumerate}

We now illustrate issue (I) by expanding two common choices of metrics applied to the full embedding $\mathbf{e}$.

\noindent\textbf{Cosine similarity (normalization effect).}
\begin{equation}
\label{eq:cosine_entanglement}
    \cos(\mathbf{e}_i,\mathbf{e}_j)=
    \frac{\mathbf{c}_i\cdot \mathbf{c}_j + \mathbf{z}_i\cdot \mathbf{z}_j}
    {\sqrt{\|\mathbf{c}_i\|^2+\|\mathbf{z}_i\|^2}\;
     \sqrt{\|\mathbf{c}_j\|^2+\|\mathbf{z}_j\|^2}}
\end{equation}
When $\|\mathbf{z}\|\gg \|\mathbf{c}\|$, the denominator is dominated by the nuisance component, so high cosine similarity largely reflects ``same topic'' rather than the intended construct (e.g., ``novel conceptual contribution'').

\noindent\textbf{Euclidean distance (additive decomposition issue).}
\begin{equation}
\label{eq:l2_entanglement}
    \|\mathbf{e}_i-\mathbf{e}_j\|^2
    = \|\mathbf{c}_i-\mathbf{c}_j\|^2 + \|\mathbf{z}_i-\mathbf{z}_j\|^2
\end{equation}
Because the contributions of $\Delta \mathbf{c}$ and $\Delta \mathbf{z}$ are purely additive, a large distance can arise from genuine conceptual deviation ($\Delta C$) or from nuisance shifts such as subfield jargon or topical drift ($\Delta Z$), and the metric alone does not identify which source drives the score.

\subsection{Strategies for Alleviation}
Because perfect disentanglement from observational text is not guaranteed without additional structure or assumptions, we outline a practical, multi-level strategy that intervenes at three points in the pipeline: (i) the input text, (ii) the learned representation, and (iii) the scoring function. The goal is to approximate measurement validity by reducing dependence on nuisance factors $Z$ while preserving information relevant to the construct $C$.

\paragraph{Level 1: Input disentanglement (preprocessing $D$).}
The most direct intervention is to reduce the dependence of the observed document $D$ on nuisance factors before representation learning. Concretely, we transform $D \rightarrow D'$ to preserve construct-relevant content while attenuating nuisance variation. \textbf{Targeted information extraction (LLM-based):} rather than embedding the full document, one can use an LLM as a constrained extractor to retain only spans that are logically relevant to the construct definition. For example, when measuring \textit{mission hybridity}, one can extract mission statements and strategic objectives while discarding biographical details, formatting artifacts, or boilerplate that primarily contribute to $Z$. \textbf{Stylistic normalization:} because embeddings are sensitive to length, register, and complexity, inputs can be normalized to a canonical form (e.g., comparable length, standardized structure, or neutral tone) to reduce stylistic nuisance components (e.g., $Z_{\text{style}}$). \textbf{Entity masking (anonymization):} when the construct is conceptual (e.g., \textit{policy stance}) but nuisance variation includes entity identity (e.g., organization names, demographic markers), embedding models may rely on entity co-occurrence rather than the intended abstract relation. Replacing named entities with placeholders (e.g., \texttt{[ORG]}, \texttt{[PERSON]}) can reduce entity-driven shortcuts and encourage reliance on structural and contextual cues.

\paragraph{Level 2: Representation disentanglement (learning in the $\mathbf{e}$ space).}
If the representation itself is entangled, we can impose objectives that explicitly suppress $Z$ while preserving $C$. \textbf{Supervised/adversarial removal:} when labels for nuisance factors are available (e.g., topic IDs, author attributes, domain indicators), adversarial learning can discourage encodings of $Z$ in $\mathbf{e}$ (e.g., Adversarial Removal; \citealp{elazar2018adversarial}), and projection-based methods can iteratively identify and remove subspaces predictive of $Z$ (e.g., Iterative Nullspace Projection; \citealp{ravfogel2020}). \textbf{Contrastive learning:} when one can construct paired examples that preserve $C$ while varying $Z$, contrastive objectives can encourage embeddings to align with construct-invariant features rather than nuisance variation.

\paragraph{Level 3: Methodological disentanglement (designing the scoring function $f$).}
When residual entanglement remains, the measurement function itself can be designed to cancel nuisance contributions. \textbf{Counterfactual neutralization:} instead of reporting a static score $f(\mathbf{e}_{\text{obs}})$, define measurement as a differential relative to a baseline embedding:
\begin{equation}
\label{eq:counterfactual_neutralization}
\hat{C} = f(\mathbf{e}_{\text{obs}}) - f(\mathbf{e}_{\text{base}}),
\end{equation}
where $\mathbf{e}_{\text{base}}$ is the embedding of a counterfactual ``neutralized'' version of the same text in which the construct signal is minimized (e.g., rewriting to remove stance/novelty claims while preserving topical content). The subtraction aims to remove the baseline contribution attributable to nuisance factors $Z$, leaving a score more sensitive to the construct-relevant component.
\section{The Construct Validity Protocol (CVP)}

Construct validity is not merely a quality check; it is a prerequisite for identification. In both social science and machine learning, isolating a target factor from nuisance variation requires additional structure (e.g., design constraints or validation evidence), much like causal representation learning requires structural assumptions to separate latent causes from noise \citep{locatello2019,scholkopf2021}. For embedding-based measurement, the core requirement is that variation in the proxy $\widehat{C}$ is attributable to the intended construct $C$, rather than superficial regularities such as topic, style, or length ($Z$) \citep{grimmer2013text,adcock2001measurement}. Without this form of \emph{measurement identification}, downstream inferences are fragile: we cannot interpret effects of a variable that has not been credibly isolated. We therefore propose the \textit{Construct Validity Protocol (CVP)} as a standard operating procedure that moves from an abstract concept to a testable measurement instrument.

\subsection{Overview}
CVP has three phases. Phase~1 establishes a construct specification and qualitative alignment. Phase~2 emphasizes instrument design choices that reduce entanglement with nuisance factors. Phase~3 reports a structured validity suite---a \textbf{Validity Card}---that documents stability and quantitative evidence (convergent, discriminant/incremental, and criterion-related checks).

\subsection{Phase 1: Conceptualization}
Before any modeling, the researcher should establish \textbf{face and content validity} through expert scrutiny \citep{adcock2001measurement,devellis2016scale,lawshe1975,haynes1995}. This phase clarifies what the construct is \emph{and is not}, and what nuisance dimensions must be neutralized.

\paragraph{Deliverables.}
\noindent\textbf{Construct map:} Provide a concise definition of $C$ and enumerate the primary nuisance dimensions $Z$ (e.g., domain/topic vocabulary, style/register, length, venue/time).
\textbf{Facet blueprint:} Specify a short list of facets of $C$ (inclusion/exclusion criteria) and the types of textual evidence expected for each facet.
\textbf{Exemplar set:} Curate a three-tier anchor set of known high-$C$, low-$C$, and borderline/mid-spectrum documents (``known-groups'' anchors) that are broadly agreed upon within the research community; borderline anchors are used to probe facet boundaries and decision rules.
\paragraph{Face-validity audit.}
Run the full measurement pipeline on the exemplar set and inspect the highest- and lowest-scoring cases, then examine borderline anchors to probe decision boundaries. Independent domain experts should judge whether the ranking is plausible and, crucially, which cues appear to drive the scores. Systematic disagreements (including non-ideal false positives and false negatives on the anchors) indicate construct--measure misalignment and should trigger revisions before quantitative validation.

\subsection{Phase 2: Operationalization} 
Rather than defaulting to black-box similarity scores, Phase~2 focuses on designing instruments that reduce entanglement with $Z$ and better isolate $C$. Section~3 outlines concrete intervention points (input preprocessing, representation learning, and scoring-function design), which can be instantiated as task-specific measurement instruments.

\subsection{Phase 3: The Validity Suite }
Validity is not a single statistic but an argument supported by multiple, conceptually distinct sources of evidence \citep{cronbach1955,campbell1959,messick1995,aera2014standards}. This is especially important for embedding-based measures, where instability and nuisance leakage are common \citep{grimmer2013text,gentzkow2019text}. We therefore recommend reporting a structured \textbf{Validity Card} that documents (at minimum) stability and discriminant/incremental evidence, and (when feasible) convergent and criterion-related evidence.\footnote{Measurement theory distinguishes multiple families of evidence, including content/face evidence, reliability/stability, convergent and discriminant evidence (external structure), and criterion-related evidence \citep{campbell1959,messick1995,aera2014standards}. We prioritize \emph{(i) stability} as a prerequisite and \emph{(ii) discriminant + incremental validity} as the most diagnostic for ruling out topic/style surrogacy in embedding pipelines; known-groups and criterion-related checks are recommended complements.}

\paragraph{Validity Card 1: Reliability / Stability (prerequisite).}
Even deterministic encoders can yield unstable measures due to preprocessing choices, chunking, or prompt/instrument sensitivity. If small, defensible perturbations materially change the resulting variable, downstream validity claims become specification-dependent \citep{koo2016icc,nunnally1994}.

\noindent\textbf{Recommended perturbations.} Swap between close encoder variants; paraphrase prompts; vary neutralization prompts; vary embedding aggregation rules.

\noindent\textbf{What to report.} (i) a clearly enumerated perturbation set and the number of variants per document; (ii) \emph{absolute-agreement} reliability via ICC (report the ICC form, e.g., ICC(2,1) or ICC(2,$k$)) \citep{shrout1979icc,koo2016icc}. As a rule of thumb, ICC $\ge 0.75$ is often interpreted as ``good'' and ICC $\ge 0.90$ as ``excellent'' \citep{koo2016icc}.\footnote{Pearson correlation can be misleading for stability because it is invariant to shifts and rescalings. ICC for absolute agreement penalizes systematic shifts in level across perturbations.}

\paragraph{Validity Card 2: Convergent validity (agreement with an independent measure).}
Does $\widehat{C}_i$ correlate with an independent measure of the same construct (e.g., expert ratings using a rubric that operationalizes $C$)? This follows multitrait--multimethod logic: measures of the same trait should agree more than measures of different traits \citep{campbell1959}.

\begin{equation}
C^{gold}_{i} = \alpha + \beta_{conv}\,\widehat{C}_{i} + \epsilon_i
\end{equation}

\noindent\textbf{What to report.} (i) reliability of $C^{gold}$ (e.g., ICC or Krippendorff's $\alpha$), since convergent evidence is bounded by gold reliability \citep{spearman1904,koo2016icc}; (ii) effect size (standardized $\beta_{conv}$ or correlation) with confidence interval and a diagnostic plot; and (iii) if possible, out-of-sample convergence via a held-out gold set or cross-validation \citep{hastie2009elements}. Effects around 0.10 are often described as small, 0.30 moderate, and 0.50+ large \citep{cohen1988statistical}.

\paragraph{Validity Card 3: Discriminant + incremental validity (primary diagnostic).}
A central threat is that $\widehat{C}$ becomes a proxy for topic, style, venue, or time rather than the intended construct \citep{grimmer2013text,gentzkow2019text}. Discriminant validity asks whether $\widehat{C}$ is distinct from nuisance dimensions. Incremental validity asks whether $\widehat{C}$ adds explanatory power beyond nuisances for an \emph{external validation label} $L_i$ that directly operationalizes $C$ (often human-coded).

\noindent\textbf{Step 1 (discriminant).} Test whether $Z$ mechanically explains $\widehat{C}$:
\begin{equation}
\widehat{C}_{i} = \alpha + \sum_{k} \gamma_k Z_{k,i} + \epsilon_i.
\end{equation}

\noindent\textbf{Step 2 (incremental).} Test whether $\widehat{C}$ adds signal for an external validation label $L_i$ beyond $Z$:
\begin{align}
L_i &= \alpha + \sum_k \delta_k Z_{k,i} + u_i, \\
L_i &= \alpha + \beta_{inc}\,\widehat{C}_i + \sum_k \delta_k Z_{k,i} + u_i.
\end{align}

\noindent\textbf{What to report.} Pre-specify nuisance blocks (topic proxies, length/style measures, venue/field/year fixed effects) and report (i) predictability of $\widehat{C}$ from $Z$ (e.g., cross-validated $R^2$) with block-wise contributions; and (ii) incremental evidence via standardized $\beta_{inc}$ with confidence interval and out-of-sample $\Delta R^2$ from adding $\widehat{C}$ to the nuisance-only model \citep{hastie2009elements,hunsley2003incremental}. High predictability of $\widehat{C}$ from topic/FE blocks is a warning sign of surrogacy; instability of $\beta_{inc}$ across reasonable nuisance sets is a red flag \citep{grimmer2013text,cohen1988statistical}.

\paragraph{Validity Card 4: Known-groups validity (optional).}
Using the Phase~1 exemplar set, test whether the measure separates pre-specified high-$C$ versus low-$C$ groups:
\begin{equation}
\widehat{C}_{i} = \alpha + \tau\,\mathbb{1}\{i\in \text{High-}C\} + \epsilon_i.
\end{equation}
\noindent\textbf{What to report.} Separation effect (e.g., Cohen's $d$ with confidence interval) and a distribution plot (e.g., ECDF). As a rule of thumb, $d\approx0.2$ is small, $0.5$ medium, and $0.8$ large \citep{cohen1988statistical}.

\paragraph{Validity Card 5: Criterion-related / predictive evidence (complement).}
If $C$ is substantively meaningful, $\widehat{C}$ should relate to downstream outcomes in the theorized direction. For validity testing, prioritize interpretability and attribution: a linear specification provides a conservative baseline that tests whether $\widehat{C}$ has independent signal beyond strong nuisance controls.

\begin{equation}
Y_{t+k, i} = \alpha + \beta_{pred}\,\widehat{C}_{i} + \mathbf{\Gamma}\text{Controls}_i + \epsilon_i.
\end{equation}

\noindent\textbf{What to report.} Standardized $\beta_{pred}$ with confidence interval and the incremental explanatory power from adding $\widehat{C}$ beyond strong nuisance controls. When possible, include a negative-control outcome $Y_{\text{placebo}}$ that should not be predicted by $C$; strong prediction of implausible outcomes suggests dataset artifacts \citep{lipsitch2010negative}.

\subsection{Worked Example}
\label{sec:cvp_goemotions}
To demonstrate that CVP is operational, we executed it on GoEmotions \citep{demszky-etal-2020-goemotions} (43{,}410 train / 5{,}426 dev / 5{,}427 test). GoEmotions is vector-valued; for brevity we instantiate one coordinate (\emph{gratitude}) with $L_i=1$ iff the human label set contains gratitude. We focus on Cards~1 and~3 since they directly test stability and nuisance entanglement; the remaining cards typically require additional independent instruments or external outcomes beyond GoEmotions.

\paragraph{Card 1.}
We instantiate the recommended perturbations by swapping close encoders
(\texttt{all-MiniLM-L6-v2} vs \texttt{all-MiniLM-L12-v2}),
varying aggregation (mean vs CLS pooling), and applying a simple style-normalization
(original vs lowercased, punctuation-stripped text), yielding $k=8$ variants
(2 encoders $\times$ 2 pooling rules $\times$ 2 normalization conditions) on $n=2000$
test examples. Each variant yields a reasonable proxy (AUC range 0.9407--0.9662). Treating variants as ``raters,'' absolute-agreement reliability is ICC(2,1)=0.8467 and ICC(2,$k$)=0.9779 \citep{shrout1979icc,koo2016icc}.

\paragraph{Card 3.}
Let $\widehat{C}_i=p(L_i=1\mid x_i)$ be a concrete proxy from sentence embeddings \citep{reimers2019sentencebert}. We form nuisance blocks $Z$ from (i) length/style features (token/character length, ``!''/``?'' counts, uppercase ratio) and (ii) a topic block (TF--IDF + SVD). Discriminant regression (Eq.~(6)) shows $\widehat{C}$ is largely nuisance-recoverable ($R^2{=}0.0245$ from length/style only; $0.7762$ from topic only; $0.7768$ from full $Z$). Incremental prediction (Eq.~(7)--(8)) improves from AUC 0.9658 ($Z{\to}L$) to 0.9831 ($Z{+}\widehat{C}{\to}L$), with $\beta_{\mathrm{inc}}{>}0$.

\paragraph{Anchors and errors (high/low/borderline).}
High-$C$: ``Thanks!!'', ``Thank you!'', ``Thanks''; Low-$C$: ``No I'm not'', ``No no she was [NAME]'', ``Why? I love it.'' Borderline: ``Thank [NAME] for the person walking down the sidewalk\ldots'', ``Did you kill the shark\ldots Thanks [NAME]!''; a salient FP is ``Yeah thank you you ungrateful bast\ldots'', and FNs include ``That's because [NAME] thinks he's the man\ldots'' and ``Daily would be great, but\ldots work\ldots''.
\section{Forensic Analysis: The Jangle Fallacy in Practice}
\label{sec:forensic}

To assess how often social NLP work substantiates the leap from an embedding-based proxy to a theoretically valid measure, we conducted a forensic coding of 17 influential papers (2020--2025) in ACL/NAACL/EMNLP that propose computational metrics for abstract social constructs. Our goal is not to single out individual contributions, but to evaluate whether the literature, in aggregate, separates (i) an \emph{empirical instrument} (a scoring function over text representations) from (ii) the \emph{theoretical construct} it claims to measure. Following standard measurement theory, we coded each paper on seven dimensions of validity and identification: construct definition, face/content evidence, reliability/stability, convergent validity, discriminant validity, predictive/criterion validity, and handling of confounding covariates (see Appendix~A for coding criteria).

\paragraph{Results.}
Table~\ref{tab:validity_counts} summarizes the distribution of evidence. Two patterns stand out. First, many papers provide \emph{conceptual framing} and basic plausibility checks: construct definition is frequently articulated (Construct Validity: 10 Yes, 7 Partial, 0 No), and most papers provide some face/content evidence (6 Yes, 11 Partial, 0 No), often in the form of annotation guidelines, curated examples, or qualitative sanity checks. Second, the evidence most diagnostic for \emph{measurement identification} is comparatively scarce. Convergent validity is rarely established against an independent instrument at the same unit of analysis (1 Yes, 12 Partial, 4 No). Discriminant validity is never fully demonstrated in the strict sense of ruling out nuisance surrogacy (0 Yes, 11 Partial, 6 No). Predictive/criterion validity under an external-outcome criterion is uncommon (1 Yes, 3 Partial, 13 No). Finally, none of the sampled papers uses causal identification methods to isolate the construct from confounds; confounding is handled primarily through heuristics or associational controls (0 Yes, 14 Partial, 3 No). Reliability/stability is the most consistently reported quantitative check (11 Yes, 4 Partial, 2 No), reflecting the routine reporting of inter-annotator agreement and, less commonly, robustness to perturbations.

\paragraph{Interpretation.}
This evidence profile is consistent with the paper's central mechanism. When a measure is implemented as a geometric function over an entangled representation, three threats dominate: (i) the score may track nuisance structure such as topic, style, length, or prompt phrasing; (ii) the mapping from geometry to construct is underdetermined without validation (metric indeterminacy); and (iii) observed associations can be driven by confounding rather than construct variation. The forensic table suggests that current practice more often documents that a proxy is \emph{defined} and \emph{stable enough to compute} than that it \emph{isolates} the intended construct.

More concretely, ``Partial'' evidence in the sample frequently takes forms that are compatible with the proxy presumption. Convergent validity is often operationalized as triangulation with related proxies, alignment with expectations, or comparisons to adjacent tasks---useful checks, but weaker than agreement with an independent validated instrument. Discriminant validity is commonly addressed through design constraints (e.g., topic balancing, prompt standardization) or qualitative discussion, rather than explicit tests showing the score is not explained by nuisance variables \emph{with respect to nuisance confounds}. Confounding is typically handled via filtering, matching, or including covariates in predictive models, which controls associationally but does not support identification without additional assumptions.

\paragraph{Implication.}
These gaps matter for cumulative science. If the literature rarely demonstrates discriminant validity and confound isolation, then two papers that both report a measure of ``bias'' or ``ideology'' can plausibly be tracking different mixtures of construct and nuisance variance. In that regime, results may appear to cumulate under a shared construct label while the underlying variables are not commensurate---a textbook setting for the \emph{jangle fallacy} \citep{kelley1927interpretation}. The forensic results therefore support the paper's broader claim: downstream utility and qualitative plausibility are often treated as sufficient, even though the evidence most necessary to distinguish construct variance from method variance remains limited. This motivates the Construct Validity Protocol in Section~4 as a minimum reporting standard for embedding-based social measures.

\begin{table}
  \centering
  \setlength{\tabcolsep}{4.2pt}
  \caption{Validity evidence across 17 social measurement papers (2020--2025). ``Yes'' indicates rigorous adherence to measurement standards; ``Partial'' indicates heuristic or indirect checks; ``No'' indicates that the dimension is not substantively evaluated.}
  \begin{tabular}{lccc}
    \hline
    \textbf{Dimension} & \textbf{YES} & \textbf{PARTIAL} & \textbf{NO} \\
    \hline
    Construct Validity                & 10 & 7  & 0  \\
    Face/Content Validity             & 6  & 11 & 0  \\
    Reliability / Stability           & 11 & 4  & 2  \\
    Convergent Validity               & 1  & 12 & 4  \\
    Discriminant Validity             & 0  & 11 & 6  \\
    Predictive Validity               & 1  & 3  & 13 \\
    Handling Confounders              & 0  & 14 & 3  \\
    \hline
  \end{tabular}
  \label{tab:validity_counts}
\end{table}
\section{Alternative Views}

\paragraph{Objection 1: ``Post-hoc Correlation is Sufficient.''}
A common counter-argument is that if a proxy correlates with a human label, disentanglement is unnecessary. We disagree. A high correlation coefficient (e.g., $r=0.7$) can be driven entirely by a confounding variable $Z$. For example, a ``Toxicity'' classifier might correlate well with human labels simply because both humans and models flag AAVE dialect as toxic \cite{zhou2022}. Without counterfactual validity (proving the score changes only when $C$ changes, not just when $Z$ changes), the metric is a biased estimator of the construct.

More broadly, learned-proxy methodology shows that post-hoc correlation is insufficient for testing causal theories with learned proxies \citep{knox2022learnedproxies}.

\paragraph{Objection 2: ``Large Scale Solves This.''}
Another view posits that sufficiently large models (e.g., GPT-4) implicitly understand the difference between concepts and nuisance variables. While LLMs are powerful generators, their embeddings remain entangled representations of the training distribution. As demonstrated by the oracle encoder fallacy, a perfect encoder preserves \textit{all} information, including the noise. Scale improves the \textit{fidelity} of the embedding, but it does not automatically perform the \textit{causal abstraction} required to separate $C$ from $Z$. Validity requires active methodological intervention, not just passive scaling.
\section{Conclusion}

The migration of NLP from engineering benchmarks to Computational Social Science demands a parallel maturation in how we \emph{measure}. We have argued that much of today's construct-oriented work implicitly relies on ``measurement by renaming''---treating convenient geometric heuristics (e.g., cosine distance in embedding space) as if they were identified measures of latent social constructs. By making the measurement problem explicit in a simple data-generating view, $D = G(C,Z)$, we show why this practice is fragile: unsupervised representations generally encode mixtures of the target construct ($C$) and nuisance attributes ($Z$) such as topic, style, venue, time, and authorship. In this setting, raw similarity metrics are not guaranteed to isolate $C$ and can instead track pipeline-induced variation.

To move from plausible proxies to cumulative measurement, we propose the \textbf{Construct Validity Protocol (CVP)} as a community standard. The CVP operationalizes a full pipeline: (i) \textbf{conceptualization} with explicit construct boundaries and a domain of observables, (ii) \textbf{face/content validation} through exemplar design and expert audit, (iii) \textbf{instrument design} that targets construct-relevant text while controlling confounds, and (iv) a \textbf{Validity Suite} that reports reliability/stability, convergent evidence against independent labels, discriminant and \emph{incremental} evidence beyond topic/style controls, known-groups separation, and predictive/criterion tests with falsification outcomes. Within this framework, we introduced \textbf{Counterfactual Neutralization}---using LLM-generated counterfactual rewrites to hold $Z$ fixed while varying construct-relevant content---and complementary tools such as orthogonal projection when appropriate.

Our aim is not to prohibit proxies, but to make them \emph{testable}. Accuracy is about hitting the target; construct validity is about ensuring the target is the right one. Embedding-based ``social variables'' should therefore be accompanied by transparent validity evidence---a compact \textbf{Validity Card} that documents design choices, stability checks, and the full set of diagnostics. Adopting such standards would make results comparable across papers, reduce the risk of topic/style leakage masquerading as social signal, and ultimately enable NLP-based measurement to serve as credible scientific evidence rather than convenient geometry.

\subsection*{Limitations}

Our coding reflects what is \emph{reported} in papers rather than what authors may have performed but did not document. This is a feature---because cumulative science depends on transparent validity arguments---but it also means we may undercount validity evidence that exists only in unpublished analyses, code repositories, or informal checks. Moreover, our coarse \{Yes/Partial/No\} labels compress a spectrum of practices, and several dimensions (especially predictive/criterion validity and confound isolation) depend on definitional choices about what constitutes a sufficiently external criterion or an identification strategy. Future work could improve reliability by preregistering the rubric, double-coding with independent raters, and reporting inter-rater agreement.

Second, this paper is a position and synthesis contribution: we propose the Construct Validity Protocol (CVP) as a standard for measurement identification, but we do not instantiate the full protocol end-to-end on a new dataset. Empirical case studies that apply the CVP prospectively---including controlled discriminant tests and confound-neutralization designs---are needed to quantify the practical costs, failure modes, and benefits of the protocol in real measurement pipelines, especially as LLM-based encoders and prompting practices continue to evolve rapidly.

\bibliography{custom}

\begin{thebibliography}{68}
\providecommand{\natexlab}[1]{#1}

\bibitem[{Adcock and Collier(2001)}]{adcock2001measurement}
Robert Adcock and David Collier. 2001.
\newblock Measurement validity: A shared standard for qualitative and quantitative research.
\newblock \emph{American political science review}, 95(3):529--546.

\bibitem[{{American Educational Research Association} et~al.(2014){American Educational Research Association}, {American Psychological Association}, and {National Council on Measurement in Education}}]{aera2014standards}
{American Educational Research Association}, {American Psychological Association}, and {National Council on Measurement in Education}. 2014.
\newblock \emph{Standards for Educational and Psychological Testing}.
\newblock American Educational Research Association.

\bibitem[{Aroyo and Welty(2015)}]{aroyo2015crowdtruth}
Lora Aroyo and Chris Welty. 2015.
\newblock Truth is a lie: Crowd truth and the seven myths of human annotation.
\newblock \emph{AI magazine}, 36(1):15--24.

\bibitem[{Azizov et~al.(2024)Azizov, Mujahid, AlQuabeh, Nakov, and Liang}]{azizov-etal-2024-safari}
Dilshod Azizov, Zain~Muhammad Mujahid, Hilal AlQuabeh, Preslav Nakov, and Shangsong Liang. 2024.
\newblock {SAFARI}: Cross-lingual bias and factuality detection in news media and news articles.
\newblock In \emph{Findings of the Association for Computational Linguistics: EMNLP 2024}, pages 12217--12231. Association for Computational Linguistics.

\bibitem[{Azzopardi and Moshfeghi(2025)}]{azzopardi-moshfeghi-2025-pow}
Leif Azzopardi and Yashar Moshfeghi. 2025.
\newblock {POW}: Political overton windows of large language models.
\newblock In \emph{Findings of the Association for Computational Linguistics: EMNLP 2025}, pages 24767--24773. Association for Computational Linguistics.

\bibitem[{Baly et~al.(2020)Baly, Da~San~Martino, Glass, and Nakov}]{baly-etal-2020-detect}
Ramy Baly, Giovanni Da~San~Martino, James Glass, and Preslav Nakov. 2020.
\newblock We can detect your bias: Predicting the political ideology of news articles.
\newblock In \emph{Proceedings of the 2020 Conference on Empirical Methods in Natural Language Processing}, pages 4982--4991.

\bibitem[{Bang et~al.(2024)Bang, Chen, Lee, and Fung}]{bang-etal-2024-measuring}
Yejin Bang, Delong Chen, Nayeon Lee, and Pascale Fung. 2024.
\newblock Measuring political bias in large language models: What is said and how it is said.
\newblock In \emph{Proceedings of the 62nd Annual Meeting of the Association for Computational Linguistics (Volume 1: Long Papers)}, pages 11142--11159.

\bibitem[{Bender and Friedman(2018)}]{bender-friedman-2018-datastatements}
Emily~M Bender and Batya Friedman. 2018.
\newblock Data statements for natural language processing: Toward mitigating system bias and enabling better science.
\newblock \emph{Transactions of the Association for Computational Linguistics}, 6:587--604.

\bibitem[{Bengio et~al.(2013)Bengio, Courville, and Vincent}]{bengio2013}
Yoshua Bengio, Aaron Courville, and Pascal Vincent. 2013.
\newblock Representation learning: A review and new perspectives.
\newblock \emph{IEEE transactions on pattern analysis and machine intelligence}, 35(8):1798--1828.

\bibitem[{Blodgett et~al.(2020)Blodgett, Barocas, Daum{\'e}~Iii, and Wallach}]{blodgett2020}
Su~Lin Blodgett, Solon Barocas, Hal Daum{\'e}~Iii, and Hanna Wallach. 2020.
\newblock Language (technology) is power: A critical survey of “bias” in nlp.
\newblock In \emph{Proceedings of the 58th annual meeting of the association for computational linguistics}, pages 5454--5476.

\bibitem[{Bolukbasi et~al.(2016)Bolukbasi, Chang, Zou, Saligrama, and Kalai}]{bolukbasi2016debias}
Tolga Bolukbasi, Kai-Wei Chang, James~Y Zou, Venkatesh Saligrama, and Adam~T Kalai. 2016.
\newblock Man is to computer programmer as woman is to homemaker? debiasing word embeddings.
\newblock \emph{Advances in Neural Information Processing Systems}, 29.

\bibitem[{Caliskan et~al.(2017)Caliskan, Bryson, and Narayanan}]{caliskan2017semantics}
Aylin Caliskan, Joanna~J Bryson, and Arvind Narayanan. 2017.
\newblock Semantics derived automatically from language corpora contain human-like biases.
\newblock \emph{Science}, 356(6334):183--186.

\bibitem[{Campbell and Fiske(1959)}]{campbell1959}
Donald~T Campbell and Donald~W Fiske. 1959.
\newblock Convergent and discriminant validation by the multitrait-multimethod matrix.
\newblock \emph{Psychological bulletin}, 56(2):81.

\bibitem[{Cohen(1988)}]{cohen1988statistical}
Jacob Cohen. 1988.
\newblock \emph{Statistical Power Analysis for the Behavioral Sciences}.
\newblock Lawrence Erlbaum Associates.

\bibitem[{Cronbach and Meehl(1955)}]{cronbach1955}
Lee~J Cronbach and Paul~E Meehl. 1955.
\newblock Construct validity in psychological tests.
\newblock \emph{Psychological bulletin}, 52(4):281.

\bibitem[{Davani et~al.(2022)Davani, D{\'\i}az, and Prabhakaran}]{davani2022disagreements}
Aida~Mostafazadeh Davani, Mark D{\'\i}az, and Vinodkumar Prabhakaran. 2022.
\newblock Dealing with disagreements: Looking beyond the majority vote in subjective annotations.
\newblock \emph{Transactions of the Association for Computational Linguistics}, 10:92--110.

\bibitem[{Demszky et~al.(2021)Demszky, Liu, Mancenido, Cohen, Hill, Jurafsky, and Hashimoto}]{demszky-etal-2021-measuring}
Dorottya Demszky, Jing Liu, Zid Mancenido, Julie Cohen, Heather Hill, Dan Jurafsky, and Tatsunori~B Hashimoto. 2021.
\newblock Measuring conversational uptake: A case study on student-teacher interactions.
\newblock In \emph{Proceedings of the 59th Annual Meeting of the Association for Computational Linguistics and the 11th International Joint Conference on Natural Language Processing (Volume 1: Long Papers)}, pages 1638--1653.

\bibitem[{Demszky et~al.(2020)Demszky, Movshovitz-Attias, Ko, Cowen, Nemade, and Ravi}]{demszky-etal-2020-goemotions}
Dorottya Demszky, Dana Movshovitz-Attias, Jeongwoo Ko, Alan Cowen, Gaurav Nemade, and Sujith Ravi. 2020.
\newblock Goemotions: A dataset of fine-grained emotions.
\newblock In \emph{Proceedings of the 58th annual meeting of the association for computational linguistics}, pages 4040--4054.

\bibitem[{DeVellis(2016)}]{devellis2016scale}
Robert~F. DeVellis. 2016.
\newblock \emph{Scale Development: Theory and Applications}.
\newblock Sage Publications.

\bibitem[{Egami et~al.(2022)Egami, Fong, Grimmer, Roberts, and Stewart}]{egami2022causaltexts}
Naoki Egami, Christian~J Fong, Justin Grimmer, Margaret~E Roberts, and Brandon~M Stewart. 2022.
\newblock How to make causal inferences using texts.
\newblock \emph{Science Advances}, 8(42):eabg2652.

\bibitem[{Egami et~al.(2023)Egami, Hinck, Stewart, and Wei}]{egami2023imperfectsurrogates}
Naoki Egami, Musashi Hinck, Brandon Stewart, and Hanying Wei. 2023.
\newblock Using imperfect surrogates for downstream inference: Design-based supervised learning for social science applications of large language models.
\newblock \emph{Advances in Neural Information Processing Systems}, 36:68589--68601.

\bibitem[{Elazar and Goldberg(2018)}]{elazar2018adversarial}
Yanai Elazar and Yoav Goldberg. 2018.
\newblock Adversarial removal of demographic attributes from text data.
\newblock In \emph{Proceedings of the 2018 Conference on Empirical Methods in Natural Language Processing}, pages 11--21.

\bibitem[{ElSherief et~al.(2021)ElSherief, Ziems, Muchlinski, Anupindi, Seybolt, De~Choudhury, and Yang}]{elsherief-etal-2021-latent}
Mai ElSherief, Caleb Ziems, David Muchlinski, Vaishnavi Anupindi, Jordyn Seybolt, Munmun De~Choudhury, and Diyi Yang. 2021.
\newblock Latent hatred: A benchmark for understanding implicit hate speech.
\newblock In \emph{Proceedings of the 2021 Conference on Empirical Methods in Natural Language Processing}, pages 345--363.

\bibitem[{Faulborn et~al.(2025)Faulborn, Sen, Pellert, Spitz, and Garcia}]{faulborn-etal-2025-little}
Mats Faulborn, Indira Sen, Max Pellert, Andreas Spitz, and David Garcia. 2025.
\newblock Only a little to the left: A theory-grounded measure of political bias in large language models.
\newblock In \emph{Proceedings of the 63rd Annual Meeting of the Association for Computational Linguistics (Volume 1: Long Papers)}, pages 31684--31704.

\bibitem[{Feng et~al.(2023)Feng, Park, Liu, and Tsvetkov}]{feng-etal-2023-pretraining}
Shangbin Feng, Chan~Young Park, Yuhan Liu, and Yulia Tsvetkov. 2023.
\newblock From pretraining data to language models to downstream tasks: Tracking the trails of political biases leading to unfair nlp models.
\newblock In \emph{Proceedings of the 61st Annual Meeting of the Association for Computational Linguistics (Volume 1: Long Papers)}, pages 11737--11762.

\bibitem[{Gabriel et~al.(2022)Gabriel, Hallinan, Sap, Nguyen, Roesner, Choi, and Choi}]{gabriel-etal-2022-misinfo}
Saadia Gabriel, Skyler Hallinan, Maarten Sap, Pemi Nguyen, Franziska Roesner, Eunsol Choi, and Yejin Choi. 2022.
\newblock Misinfo reaction frames: Reasoning about readers’ reactions to news headlines.
\newblock In \emph{Proceedings of the 60th Annual Meeting of the Association for Computational Linguistics (Volume 1: Long Papers)}, pages 3108--3127.

\bibitem[{Gardner et~al.(2020)Gardner, Artzi, Basmov, Berant, Bogin, Chen, Dasigi, Dua, Elazar, Gottumukkala et~al.}]{gardner2020contrast}
Matt Gardner, Yoav Artzi, Victoria Basmov, Jonathan Berant, Ben Bogin, Sihao Chen, Pradeep Dasigi, Dheeru Dua, Yanai Elazar, Ananth Gottumukkala, and 1 others. 2020.
\newblock Evaluating models’ local decision boundaries via contrast sets.
\newblock In \emph{Findings of the Association for Computational Linguistics: EMNLP 2020}, pages 1307--1323.

\bibitem[{Garg et~al.(2018)Garg, Schiebinger, Jurafsky, and Zou}]{garg2018stereotypes}
Nikhil Garg, Londa Schiebinger, Dan Jurafsky, and James Zou. 2018.
\newblock Word embeddings quantify 100 years of gender and ethnic stereotypes.
\newblock \emph{Proceedings of the National Academy of Sciences}, 115(16):E3635--E3644.

\bibitem[{Geiger et~al.(2020)Geiger, Yu, Yang, Dai, Qiu, Tang, and Huang}]{geiger2020garbage}
R~Stuart Geiger, Kevin Yu, Yanlai Yang, Mindy Dai, Jie Qiu, Rebekah Tang, and Jenny Huang. 2020.
\newblock Garbage in, garbage out? do machine learning application papers in social computing report where human-labeled training data comes from?
\newblock In \emph{Proceedings of the 2020 conference on fairness, accountability, and transparency}, pages 325--336.

\bibitem[{Gentzkow et~al.(2019)Gentzkow, Kelly, and Taddy}]{gentzkow2019text}
Matthew Gentzkow, Bryan Kelly, and Matt Taddy. 2019.
\newblock Text as data.
\newblock \emph{Journal of Economic Literature}, 57(3):535--574.

\bibitem[{Grimmer and Stewart(2013)}]{grimmer2013text}
Justin Grimmer and Brandon~M Stewart. 2013.
\newblock Text as data: The promise and pitfalls of automatic content analysis methods for political texts.
\newblock \emph{Political analysis}, 21(3):267--297.

\bibitem[{Gururangan et~al.(2018)Gururangan, Swayamdipta, Levy, Schwartz, Bowman, and Smith}]{gururangan2018annotation}
Suchin Gururangan, Swabha Swayamdipta, Omer Levy, Roy Schwartz, Samuel Bowman, and Noah~A Smith. 2018.
\newblock Annotation artifacts in natural language inference data.
\newblock In \emph{Proceedings of the 2018 Conference of the North American Chapter of the Association for Computational Linguistics: Human Language Technologies, Volume 2 (Short Papers)}, pages 107--112.

\bibitem[{Harel-Canada et~al.(2024)Harel-Canada, Zhou, Muppalla, Yildiz, Kim, Sahai, and Peng}]{harel-canada-etal-2024-measuring}
Fabrice~Y Harel-Canada, Hanyu Zhou, Sreya Muppalla, Zeynep~Senahan Yildiz, Miryung Kim, Amit Sahai, and Nanyun Peng. 2024.
\newblock Measuring psychological depth in language models.
\newblock In \emph{Proceedings of the 2024 Conference on Empirical Methods in Natural Language Processing}, pages 17162--17196.

\bibitem[{Hastie et~al.(2009)Hastie, Tibshirani, Friedman et~al.}]{hastie2009elements}
Trevor Hastie, Robert Tibshirani, Jerome Friedman, and 1 others. 2009.
\newblock The elements of statistical learning.

\bibitem[{Haynes et~al.(1995)Haynes, Richard, and Kubany}]{haynes1995}
Stephen~N Haynes, David Richard, and Edward~S Kubany. 1995.
\newblock Content validity in psychological assessment: A functional approach to concepts and methods.
\newblock \emph{Psychological assessment}, 7(3):238.

\bibitem[{Hoover et~al.(2020)Hoover, Portillo-Wightman, Yeh, Havaldar, Davani, Lin, Kennedy, Atari, Kamel, Mendlen et~al.}]{hoover2020moral}
Joe Hoover, Gwenyth Portillo-Wightman, Leigh Yeh, Shreya Havaldar, Aida~Mostafazadeh Davani, Ying Lin, Brendan Kennedy, Mohammad Atari, Zahra Kamel, Madelyn Mendlen, and 1 others. 2020.
\newblock Moral foundations twitter corpus: A collection of 35k tweets annotated for moral sentiment.
\newblock \emph{Social Psychological and Personality Science}, 11(8):1057--1071.

\bibitem[{Huang and Yang(2023)}]{cali2023}
Jing Huang and Diyi Yang. 2023.
\newblock Culturally aware natural language inference.
\newblock In \emph{Findings of the Association for Computational Linguistics: EMNLP 2023}, pages 7591--7609.

\bibitem[{Hunsley and Meyer(2003)}]{hunsley2003incremental}
John Hunsley and Gregory~J Meyer. 2003.
\newblock The incremental validity of psychological testing and assessment: conceptual, methodological, and statistical issues.
\newblock \emph{Psychological assessment}, 15(4):446.

\bibitem[{Ils et~al.(2021)Ils, Liu, Grunow, and Eger}]{ils-etal-2021-united}
Alexandra Ils, Dan Liu, Daniela Grunow, and Steffen Eger. 2021.
\newblock Changes in european solidarity before and during covid-19: Evidence from a large crowd-and expert-annotated twitter dataset.
\newblock In \emph{Proceedings of the 59th Annual Meeting of the Association for Computational Linguistics and the 11th International Joint Conference on Natural Language Processing (Volume 1: Long Papers)}, pages 1623--1637.

\bibitem[{Kelley(1927)}]{kelley1927interpretation}
Truman~Lee Kelley. 1927.
\newblock \emph{Interpretation of educational measurements}.
\newblock World Book Company.

\bibitem[{Knox et~al.(2022)Knox, Lucas, and Cho}]{knox2022learnedproxies}
Dean Knox, Christopher Lucas, and Wendy K~Tam Cho. 2022.
\newblock Testing causal theories with learned proxies.
\newblock \emph{Annual Review of Political Science}, 25:419--441.

\bibitem[{Koo and Li(2016)}]{koo2016icc}
Terry~K Koo and Mae~Y Li. 2016.
\newblock A guideline of selecting and reporting intraclass correlation coefficients for reliability research.
\newblock \emph{Journal of chiropractic medicine}, 15(2):155--163.

\bibitem[{Kozlowski et~al.(2019)Kozlowski, Taddy, and Evans}]{kozlowski2019geometry}
Austin~C Kozlowski, Matt Taddy, and James~A Evans. 2019.
\newblock The geometry of culture: Analyzing the meanings of class through word embeddings.
\newblock \emph{American Sociological Review}, 84(5):905--949.

\bibitem[{Lawshe(1975)}]{lawshe1975}
Charles~H Lawshe. 1975.
\newblock A quantitative approach to content validity.
\newblock \emph{Personnel psychology}, 28(4).

\bibitem[{Lee et~al.(2024)Lee, Cho, Cho, Jin, Lee, and Song}]{lee-etal-2024-icscore}
Junha Lee, Jaeshin Cho, Youngjin Cho, Hyewon Jin, Hyemin Lee, and Min Song. 2024.
\newblock \href {https://neurips.cc/virtual/2024/107877} {{ICScore}: Metrics for evaluating interestingness and creativity of stories}.
\newblock Poster, NeurIPS 2024 Workshop: Statistical Frontiers in LLMs and Foundation Models.

\bibitem[{Lipsitch et~al.(2010)Lipsitch, Tchetgen, and Cohen}]{lipsitch2010negative}
Marc Lipsitch, Eric~Tchetgen Tchetgen, and Ted Cohen. 2010.
\newblock Negative controls: a tool for detecting confounding and bias in observational studies.
\newblock \emph{Epidemiology}, 21(3):383--388.

\bibitem[{Liu et~al.(2022)Liu, Zhang, Wegsman, Beauchamp, and Wang}]{liu-etal-2022-politics}
Yujian Liu, Xinliang~Frederick Zhang, David Wegsman, Nicholas Beauchamp, and Lu~Wang. 2022.
\newblock Politics: Pretraining with same-story article comparison for ideology prediction and stance detection.
\newblock In \emph{Findings of the Association for Computational Linguistics: NAACL 2022}, pages 1354--1374.

\bibitem[{Locatello et~al.(2019)Locatello, Bauer, Lucic, Raetsch, Gelly, Sch{\"o}lkopf, and Bachem}]{locatello2019}
Francesco Locatello, Stefan Bauer, Mario Lucic, Gunnar Raetsch, Sylvain Gelly, Bernhard Sch{\"o}lkopf, and Olivier Bachem. 2019.
\newblock Challenging common assumptions in the unsupervised learning of disentangled representations.
\newblock In \emph{international conference on machine learning}, pages 4114--4124. PMLR.

\bibitem[{May et~al.(2019)May, Wang, Bordia, Bowman, and Rudinger}]{may2019seat}
Chandler May, Alex Wang, Shikha Bordia, Samuel Bowman, and Rachel Rudinger. 2019.
\newblock On measuring social biases in sentence encoders.
\newblock In \emph{Proceedings of the 2019 Conference of the North American Chapter of the Association for Computational Linguistics: Human Language Technologies, Volume 1 (Long and Short Papers)}, pages 622--628.

\bibitem[{McCoy et~al.(2019)McCoy, Pavlick, and Linzen}]{mccoy2019right}
R~Thomas McCoy, Ellie Pavlick, and Tal Linzen. 2019.
\newblock Right for the wrong reasons: Diagnosing syntactic heuristics in natural language inference.
\newblock In \emph{Proceedings of the 57th annual meeting of the association for computational linguistics}, pages 3428--3448.

\bibitem[{Merrill et~al.(2024)Merrill, Smith, and Elazar}]{merrill-etal-2024-evaluating}
William Merrill, Noah~A Smith, and Yanai Elazar. 2024.
\newblock Evaluating n-gram novelty of language models using rusty-dawg.
\newblock In \emph{Proceedings of the 2024 Conference on Empirical Methods in Natural Language Processing}, pages 14459--14473.

\bibitem[{Messick(1995)}]{messick1995}
Samuel Messick. 1995.
\newblock Validity of psychological assessment: Validation of inferences from persons' responses and performances as scientific inquiry into score meaning.
\newblock \emph{American psychologist}, 50(9):741.

\bibitem[{Nunnally and Bernstein(1994)}]{nunnally1994}
J.C. Nunnally and I.H. Bernstein. 1994.
\newblock \href {https://books.google.com.sg/books?id=r0fuAAAAMAAJ} {\emph{Psychometric Theory}}.
\newblock Number no. 972 in McGraw-Hill series in psychology. McGraw-Hill Companies,Incorporated.

\bibitem[{Pavlick and Kwiatkowski(2019)}]{pavlick-kwiatkowski-2019-disagreements}
Ellie Pavlick and Tom Kwiatkowski. 2019.
\newblock Inherent disagreements in human textual inferences.
\newblock \emph{Transactions of the Association for Computational Linguistics}, 7:677--694.

\bibitem[{Plank(2022)}]{plank-2022-labelvariation}
Barbara Plank. 2022.
\newblock The “problem” of human label variation: On ground truth in data, modeling and evaluation.
\newblock In \emph{Proceedings of the 2022 conference on Empirical Methods in Natural Language Processing}, pages 10671--10682.

\bibitem[{Poliak et~al.(2018)Poliak, Naradowsky, Haldar, Rudinger, and Van~Durme}]{poliak2018hypothesis}
Adam Poliak, Jason Naradowsky, Aparajita Haldar, Rachel Rudinger, and Benjamin Van~Durme. 2018.
\newblock Hypothesis only baselines in natural language inference.
\newblock In \emph{Proceedings of the seventh joint conference on lexical and computational semantics}, pages 180--191.

\bibitem[{Ravfogel et~al.(2020)Ravfogel, Elazar, Gonen, Twiton, and Goldberg}]{ravfogel2020}
Shauli Ravfogel, Yanai Elazar, Hila Gonen, Michael Twiton, and Yoav Goldberg. 2020.
\newblock Null it out: Guarding protected attributes by iterative nullspace projection.
\newblock In \emph{Proceedings of the 58th annual meeting of the association for computational linguistics}, pages 7237--7256.

\bibitem[{Reimers and Gurevych(2019)}]{reimers2019sentencebert}
Nils Reimers and Iryna Gurevych. 2019.
\newblock Sentence-bert: Sentence embeddings using siamese bert-networks.
\newblock In \emph{Proceedings of the 2019 Conference on Empirical Methods in Natural Language Processing and the 9th International Joint Conference on Natural Language Processing (EMNLP-IJCNLP)}, pages 3982--3992.

\bibitem[{Ribeiro et~al.(2020)Ribeiro, Wu, Guestrin, and Singh}]{ribeiro2020checklist}
Marco~Tulio Ribeiro, Tongshuang Wu, Carlos Guestrin, and Sameer Singh. 2020.
\newblock Beyond accuracy: Behavioral testing of nlp models with checklist.
\newblock In \emph{Proceedings of the 58th annual meeting of the association for computational linguistics}, pages 4902--4912.

\bibitem[{Sap et~al.(2019)Sap, Card, Gabriel, Choi, and Smith}]{sap2019risk}
Maarten Sap, Dallas Card, Saadia Gabriel, Yejin Choi, and Noah~A Smith. 2019.
\newblock The risk of racial bias in hate speech detection.
\newblock In \emph{Proceedings of the 57th annual meeting of the Association for Computational Linguistics}, pages 1668--1678.

\bibitem[{Sch{\"o}lkopf et~al.(2021)Sch{\"o}lkopf, Locatello, Bauer, Ke, Kalchbrenner, Goyal, and Bengio}]{scholkopf2021}
Bernhard Sch{\"o}lkopf, Francesco Locatello, Stefan Bauer, Nan~Rosemary Ke, Nal Kalchbrenner, Anirudh Goyal, and Yoshua Bengio. 2021.
\newblock Toward causal representation learning.
\newblock \emph{Proceedings of the IEEE}, 109(5):612--634.

\bibitem[{Shrout and Fleiss(1979)}]{shrout1979icc}
Patrick~E Shrout and Joseph~L Fleiss. 1979.
\newblock Intraclass correlations: uses in assessing rater reliability.
\newblock \emph{Psychological bulletin}, 86(2):420.

\bibitem[{Sinno et~al.(2022)Sinno, Oviedo, Atwell, Alikhani, and Li}]{sinno-etal-2022-political}
Barea Sinno, Bernardo Oviedo, Katherine Atwell, Malihe Alikhani, and Junyi~Jessy Li. 2022.
\newblock Political ideology and polarization: A multi-dimensional approach.
\newblock In \emph{Proceedings of the 2022 Conference of the North American Chapter of the Association for Computational Linguistics: Human Language Technologies}, pages 231--243.

\bibitem[{Sky et~al.(2023)Sky, Saakyan, Li, Yu, and Muresan}]{ch-wang-etal-2023-sociocultural}
CH-Wang Sky, Arkadiy Saakyan, Oliver Li, Zhou Yu, and Smaranda Muresan. 2023.
\newblock Sociocultural norm similarities and differences via situational alignment and explainable textual entailment.
\newblock In \emph{Proceedings of the 2023 Conference on Empirical Methods in Natural Language Processing}, pages 3548--3564.

\bibitem[{Spearman(1904)}]{spearman1904}
Charles Spearman. 1904.
\newblock The proof and measurement of association between two things.
\newblock \emph{The American Journal of Psychology}, 15(1):72--101.

\bibitem[{Vidgen et~al.(2021)Vidgen, Thrush, Talat, and Kiela}]{vidgen-etal-2021-learning}
Bertie Vidgen, Tristan Thrush, Zeerak Talat, and Douwe Kiela. 2021.
\newblock Learning from the worst: Dynamically generated datasets to improve online hate detection.
\newblock In \emph{Proceedings of the 59th annual meeting of the Association for Computational Linguistics and the 11th international joint conference on natural language processing (volume 1: long papers)}, pages 1667--1682.

\bibitem[{Vijjini et~al.(2024)Vijjini, Menon, Fu, Srivastava, and Chaturvedi}]{socialgaze2024}
Anvesh~Rao Vijjini, Rakesh~R Menon, Jiayi Fu, Shashank Srivastava, and Snigdha Chaturvedi. 2024.
\newblock Socialgaze: Improving the integration of human social norms in large language models.
\newblock In \emph{Findings of the Association for Computational Linguistics: EMNLP 2024}, pages 16487--16506.

\bibitem[{Zhou et~al.(2022)Zhou, Ethayarajh, Card, and Jurafsky}]{zhou2022}
Kaitlyn Zhou, Kawin Ethayarajh, Dallas Card, and Dan Jurafsky. 2022.
\newblock Problems with cosine as a measure of embedding similarity for high frequency words.
\newblock In \emph{Proceedings of the 60th Annual Meeting of the Association for Computational Linguistics (Volume 2: Short Papers)}, pages 401--423.

\end{thebibliography}

\appendix

\section{Measurement Validity Coding Scheme and Rubric}
\label{app:rubric}

This appendix specifies the coding rubric used in the forensic analysis (Appendix~\ref{app:forensic}). Each paper is evaluated on seven dimensions of measurement validity and identification. For each dimension, we assign one of three labels: \textsc{Yes}, \textsc{Partial}, or \textsc{No}.

\subsection{General Scoring Labels}
\label{app:rubric_general}

\begin{itemize}
    \item \textbf{\textsc{Yes} (High rigor).} The method meets premium social science standards. It is grounded in established theory and/or uses rigorous psychometric validation or explicit causal designs, with clear empirical evidence.
    \item \textbf{\textsc{Partial} (Medium rigor).} The method relies on operational heuristics, ``silver standards,'' or ad-hoc sanity checks. It may acknowledge threats (e.g., confounding) but addresses them with standard supervised learning, informal qualitative inspection, or design constraints rather than explicit identification.
    \item \textbf{\textsc{No} (Low rigor).} The authors provide no discussion, definition, or empirical validation for the dimension.
\end{itemize}

\subsection{Dimensions of Validity}
\label{app:rubric_dimensions}

\paragraph{Dimension 1: Target Variable Definition (Construct Validity).}
\begin{itemize}
    \item \textbf{\textsc{Yes}.} Cites a specific, pre-existing social science theory (e.g., Moral Foundations Theory) and defines the construct independently of the dataset and measurement procedure (e.g., ``We measure ideology as conceptually defined by Converse (1964) \ldots'').
    \item \textbf{\textsc{Partial}.} Defines the variable operationally, tautologically, or based primarily on dataset labels or a single chosen proxy (e.g., ``toxicity is whatever the Perspective API labels as toxic,'' or ``bias is cosine distance between vectors'').
    \item \textbf{\textsc{No}.} No clear definition is provided; the variable is treated as self-explanatory.
\end{itemize}

\paragraph{Dimension 2: Face Validity (Content Validity).}
\begin{itemize}
    \item \textbf{\textsc{Yes}.} Conducts a formal content validity study prior to deployment, such as expert-panel review of items/lexicon or a structured pilot with domain experts.
    \item \textbf{\textsc{Partial}.} Provides informal sanity checks (e.g., selected high/low scoring examples, qualitative inspection of top features/words) without a structured expert evaluation.
    \item \textbf{\textsc{No}.} No inspection of instrument content and no qualitative examples are provided.
\end{itemize}

\paragraph{Dimension 3: Reliability / Stability.}
\begin{itemize}
    \item \textbf{\textsc{Yes}.} Reports formal, chance-corrected reliability/stability metrics. For human annotation: Cohen's/Fleiss' $\kappa$, ICC, or Krippendorff's $\alpha$. For model-based instruments: test--retest reliability or rigorous robustness checks (e.g., prompt perturbations/paraphrasing) with appropriate agreement metrics.
    \item \textbf{\textsc{Partial}.} Reports weak or non-corrected metrics (e.g., raw \% agreement without chance correction), or evaluates stability only on a small/non-representative subset.
    \item \textbf{\textsc{No}.} No reliability metric or stability check is reported.
\end{itemize}

\paragraph{Dimension 4: Convergent Validity.}
\begin{itemize}
    \item \textbf{\textsc{Yes}.} Demonstrates significant association with a completely independent, external ``gold-standard'' measure of the same construct at the appropriate unit of analysis (e.g., ``Our ideology score correlates with DW-NOMINATE roll-call votes'').
    \item \textbf{\textsc{Partial}.} Correlates with a ``silver standard,'' a related-but-distinct proxy, or internal metadata that is suggestive but not a gold-standard instrument (e.g., star ratings, hashtags, outlet categories).
    \item \textbf{\textsc{No}.} No external correlation or comparison is reported.
\end{itemize}

\paragraph{Dimension 5: Discriminant Validity.}
\begin{itemize}
    \item \textbf{\textsc{Yes}.} Explicitly tests and empirically demonstrates that the measure is distinct from nuisance factors (e.g., near-zero correlation with document length, explicit topic residualization, or formal tests against style/genre/demographic/prompt effects).
    \item \textbf{\textsc{Partial}.} Acknowledges potential confounds and/or offers qualitative arguments or design constraints, but does not provide formal tests sufficient to rule out nuisance surrogacy.
    \item \textbf{\textsc{No}.} No discussion or test of discriminant validity.
\end{itemize}

\paragraph{Dimension 6: Predictive Validity.}
\begin{itemize}
    \item \textbf{\textsc{Yes}.} Predicts a real-world downstream outcome external to the annotation/task setting (e.g., ``polarization predicts future protest violence,'' ``trust predicts trading volume'').
    \item \textbf{\textsc{Partial}.} Predicts an internal proxy outcome or trivial metadata label (e.g., predicting subreddit labels, publication year, or other dataset-internal fields).
    \item \textbf{\textsc{No}.} No predictive/criterion task is reported.
\end{itemize}

\paragraph{Dimension 7: Handling Confounding Covariates.}
\begin{itemize}
    \item \textbf{\textsc{Yes}.} Uses explicit causal inference methods to isolate the construct from confounding covariates $Z$ (e.g., instrumental variables, double machine learning, propensity score matching, or rigorous residualization under stated assumptions).
    \item \textbf{\textsc{Partial}.} Uses heuristics to exclude/limit $Z$, or uses standard supervised learning that includes $Z$ as a feature (associative ``controls'' without causal identification).
    \item \textbf{\textsc{No}.} No discussion of confounding covariates; treats relationships as direct without controls.
\end{itemize}

\section{Paper-Level Forensic Coding Notes}
\label{app:forensic}
This appendix reports the paper-level coding notes underlying Table~\ref{tab:validity_counts} in the main text. We code 17 influential ACL/NAACL/EMNLP papers (2020--2025) that propose computational measures of abstract social constructs using seven dimensions: (D1) target-variable definition (construct validity), (D2) face/content validity, (D3) reliability/stability, (D4) convergent validity, (D5) discriminant validity, (D6) predictive/criterion validity, and (D7) handling confounding covariates. Each dimension is coded as \textsc{Yes}, \textsc{Partial}, or \textsc{No} following the rubric in Appendix~\ref{app:rubric}.

\subsection{Paper-Level Coding Notes (D1--D7)}
\label{app:paper_notes}

\paragraph{Paper 1: \citeauthor{demszky-etal-2020-goemotions} (ACL \citeyear{demszky-etal-2020-goemotions}) --- Fine-grained Emotions (Reddit).}
\begin{itemize}
    \item \textbf{D1 (Construct): \textsc{Partial}.} Label set motivated by psychology literature and a careful selection process, but the construct definition remains closely tied to the taxonomy rather than an independent theory-grounded construct map.
    \item \textbf{D2 (Face): \textsc{Partial}.} Presents labeled examples (e.g., example annotations), but no formal expert content-validity audit is reported.
    \item \textbf{D3 (Reliability): \textsc{Yes}.} Reports rater agreement metrics including chance-corrected statistics (e.g., Cohen's $\kappa$) alongside corroborating checks.
    \item \textbf{D4 (Convergent): \textsc{Partial}.} Shows transfer/generalization to existing emotion benchmarks; useful triangulation but not a classic external gold-standard convergence test.
    \item \textbf{D5 (Discriminant): \textsc{No}.} No explicit tests that emotion labels are distinct from nuisance factors (topic/style/length).
    \item \textbf{D6 (Predictive): \textsc{No}.} Validations focus on label prediction rather than external real-world criteria.
    \item \textbf{D7 (Confounding): \textsc{Partial}.} Identifies confounds and applies curation measures, but no causal identification strategy is used.
\end{itemize}

\paragraph{Paper 2: \citeauthor{baly-etal-2020-detect} (EMNLP \citeyear{baly-etal-2020-detect}) --- Political Bias and Factuality (Outlets + Social Profiles).}
\begin{itemize}
    \item \textbf{D1 (Construct): \textsc{Partial}.} Targets defined primarily through MBFC label categories rather than an independent theory definition.
    \item \textbf{D2 (Face): \textsc{Partial}.} Informal sanity checks (e.g., excluding ill-defined label categories), but no structured content-validity study.
    \item \textbf{D3 (Reliability): \textsc{No}.} Reliability of the inherited labels is not quantified in the paper.
    \item \textbf{D4 (Convergent): \textsc{No}.} Evaluations mainly predict the same label scheme; no external convergence at the same unit of analysis.
    \item \textbf{D5 (Discriminant): \textsc{No}.} Feature ablations are not framed as discriminant tests against nuisances.
    \item \textbf{D6 (Predictive): \textsc{No}.} No external outcome prediction beyond the dataset labels.
    \item \textbf{D7 (Confounding): \textsc{Partial}.} Practical heuristics (filters/exclusions/multi-source signals), but no causal identification methods.
\end{itemize}

\paragraph{Paper 3: \citeauthor{ils-etal-2021-united} (ACL \citeyear{ils-etal-2021-united}) --- Social Solidarity / Anti-Solidarity (Twitter).}
\begin{itemize}
    \item \textbf{D1 (Construct): \textsc{Yes}.} Provides an explicit, theory-grounded definition of solidarity.
    \item \textbf{D2 (Face): \textsc{Yes}.} Involves social-science experts in refinement/adjudication.
    \item \textbf{D3 (Reliability): \textsc{Yes}.} Reports Cohen's $\kappa$ and compares agreement across conditions.
    \item \textbf{D4 (Convergent): \textsc{No}.} No convergence test against an external solidarity index/instrument.
    \item \textbf{D5 (Discriminant): \textsc{Partial}.} Reports limited discriminant evidence (e.g., low correlation with sentiment), but no systematic nuisance-factor testing.
    \item \textbf{D6 (Predictive): \textsc{Partial}.} Correlates the construct with external indicators (e.g., COVID-19 rates) at aggregate level; suggestive but not a dedicated predictive validation.
    \item \textbf{D7 (Confounding): \textsc{Partial}.} Notes sampling and causal ambiguity; no identification strategy is implemented.
\end{itemize}

\paragraph{Paper 4: \citeauthor{demszky-etal-2021-measuring} (ACL \citeyear{demszky-etal-2021-measuring}) --- Conversational Uptake (Student--Teacher Transcripts).}
\begin{itemize}
    \item \textbf{D1 (Construct): \textsc{Yes}.} Defines uptake as a linguistic/social construct and provides an operational definition.
    \item \textbf{D2 (Face): \textsc{Partial}.} Offers qualitative plausibility checks and examples but no formal content-validity study.
    \item \textbf{D3 (Reliability): \textsc{Yes}.} Reports structured annotation and inter-rater agreement with explicit statistics; describes aggregation.
    \item \textbf{D4 (Convergent): \textsc{Yes}.} Tests alignment with indicators expected to track uptake.
    \item \textbf{D5 (Discriminant): \textsc{Partial}.} Shows uptake is not reducible to repetition/overlap, but does not fully rule out broader nuisances.
    \item \textbf{D6 (Predictive): \textsc{Yes}.} Links uptake to downstream outcomes relevant to teaching contexts (e.g., satisfaction/quality).
    \item \textbf{D7 (Confounding): \textsc{Partial}.} Addresses some topical overlap concerns but does not implement causal identification.
\end{itemize}

\paragraph{Paper 5: \citeauthor{vidgen-etal-2021-learning} (ACL \citeyear{vidgen-etal-2021-learning}) --- Online Hate (Adversarial Dataset).}
\begin{itemize}
    \item \textbf{D1 (Construct): \textsc{Yes}.} Provides an explicit definition of hate for the annotation task.
    \item \textbf{D2 (Face): \textsc{Yes}.} Emphasizes expert annotators and presents credibility signals for labeling.
    \item \textbf{D3 (Reliability): \textsc{Yes}.} Reports chance-corrected agreement (e.g., Krippendorff's $\alpha$) across rounds.
    \item \textbf{D4 (Convergent): \textsc{Partial}.} Evaluates against an external functional test suite (triangulation rather than gold-standard correlation).
    \item \textbf{D5 (Discriminant): \textsc{Partial}.} Uses hard negatives/contrastive design to reduce keyword confounds, but does not provide formal discriminant statistics.
    \item \textbf{D6 (Predictive): \textsc{No}.} Focuses on dataset/model evaluation, not external real-world outcomes.
    \item \textbf{D7 (Confounding): \textsc{Partial}.} Mitigates confounds via data design, but no causal identification strategy is used.
\end{itemize}

\paragraph{Paper 6: \citeauthor{elsherief-etal-2021-latent} (EMNLP \citeyear{elsherief-etal-2021-latent}) --- Latent Hatred / Implicit Hate (Twitter).}
\begin{itemize}
    \item \textbf{D1 (Construct): \textsc{Yes}.} Develops a typology targeting implicit hate beyond explicit slurs.
    \item \textbf{D2 (Face): \textsc{Partial}.} Taxonomy and examples support plausibility, but no formal expert panel content-validity study is reported.
    \item \textbf{D3 (Reliability): \textsc{Yes}.} Reports chance-corrected reliability (e.g., ICC, Fleiss' $\kappa$).
    \item \textbf{D4 (Convergent): \textsc{Partial}.} Compares against widely used systems; informative but not an external gold-standard construct instrument.
    \item \textbf{D5 (Discriminant): \textsc{Partial}.} Distinguishes implicit from explicit hate; limited evidence against adjacent nuisances (offensiveness/negativity).
    \item \textbf{D6 (Predictive): \textsc{No}.} Evaluations are mainly benchmarking/model performance.
    \item \textbf{D7 (Confounding): \textsc{Partial}.} Uses heuristics (e.g., filtering/keyword-related controls) to focus on the implicit construct, but not causal identification.
\end{itemize}

\paragraph{Paper 7: \citeauthor{liu-etal-2022-politics} (NAACL \citeyear{liu-etal-2022-politics}) --- Ideology and Stance (News; POLITICS pretraining).}
\begin{itemize}
    \item \textbf{D1 (Construct): \textsc{Partial}.} Ideology operationalized via outlet-level labels rather than theory-grounded definition.
    \item \textbf{D2 (Face): \textsc{Partial}.} Informal plausibility checks (e.g., attention visualization), but no structured content-validity procedure.
    \item \textbf{D3 (Reliability): \textsc{No}.} No reliability metrics reported for outlet-level ideology labels.
    \item \textbf{D4 (Convergent): \textsc{Partial}.} Uses third-party outlet ratings and tests downstream utility, but no independent political-science ideology measure at same unit.
    \item \textbf{D5 (Discriminant): \textsc{Partial}.} Limited topic-control evidence; no systematic nuisance testing.
    \item \textbf{D6 (Predictive): \textsc{No}.} Focuses on ideology prediction/benchmarks rather than external outcomes.
    \item \textbf{D7 (Confounding): \textsc{Partial}.} Uses heuristics to reduce imbalance/bias; no causal identification methods.
\end{itemize}

\paragraph{Paper 8: \citeauthor{sinno-etal-2022-political} (NAACL \citeyear{sinno-etal-2022-political}) --- Multi-dimensional Political Ideology (News Paragraphs).}
\begin{itemize}
    \item \textbf{D1 (Construct): \textsc{Yes}.} Political ideology defined with political-science grounding and operationalized across dimensions.
    \item \textbf{D2 (Face): \textsc{Yes}.} Provides interpretable examples and expert annotation credibility signals.
    \item \textbf{D3 (Reliability): \textsc{Yes}.} Reports chance-corrected agreement (Krippendorff's $\alpha$).
    \item \textbf{D4 (Convergent): \textsc{Partial}.} Triangulates with outlet-bias signals; not a direct external instrument correlation.
    \item \textbf{D5 (Discriminant): \textsc{Partial}.} Separates ideology from stance by design; no formal nuisance-confound tests beyond constraints.
    \item \textbf{D6 (Predictive): \textsc{No}.} Evaluations are classification/label prediction.
    \item \textbf{D7 (Confounding): \textsc{Partial}.} Uses design/annotation constraints (e.g., controlling for stance), but no verified causal isolation.
\end{itemize}

\paragraph{Paper 9: \citeauthor{gabriel-etal-2022-misinfo} (ACL \citeyear{gabriel-etal-2022-misinfo}) --- Misinfo Reaction Frames (Headlines).}
\begin{itemize}
    \item \textbf{D1 (Construct): \textsc{Partial}.} Defines dimensions operationally; construct remains task-specific rather than externally standardized.
    \item \textbf{D2 (Face): \textsc{Partial}.} Provides examples/descriptions; crowd annotations without expert-panel content validity.
    \item \textbf{D3 (Reliability): \textsc{Partial}.} Reports reliability for categorical judgment; limited/no parallel reliability reporting for free-text dimensions.
    \item \textbf{D4 (Convergent): \textsc{No}.} No validation against external behavioral ground truth of reactions.
    \item \textbf{D5 (Discriminant): \textsc{No}.} No explicit tests distinguishing the construct from nuisances.
    \item \textbf{D6 (Predictive): \textsc{Partial}.} Shows systematic shifts in trust ratings in a controlled setting; not real-world behavioral criterion validation.
    \item \textbf{D7 (Confounding): \textsc{Partial}.} Design mitigations exist, but no causal confound-handling pipeline.
\end{itemize}

\paragraph{Paper 10: \citeauthor{hoover2020moral} (\citeyear{hoover2020moral}) --- Moral Sentiment (MFT; Tweets).}
\begin{itemize}
    \item \textbf{D1 (Construct): \textsc{Yes}.} Categories grounded in Moral Foundations Theory.
    \item \textbf{D2 (Face): \textsc{Yes}.} Provides structured guidelines and annotator training procedures; early disagreement handling.
    \item \textbf{D3 (Reliability): \textsc{Yes}.} Reports chance-corrected agreement (e.g., Fleiss' $\kappa$, PABAK).
    \item \textbf{D4 (Convergent): \textsc{Partial}.} No correlation with an independent external MFT instrument; convergence mostly indirect.
    \item \textbf{D5 (Discriminant): \textsc{Partial}.} Conceptual separation of categories without formal nuisance-factor tests.
    \item \textbf{D6 (Predictive): \textsc{No}.} Benchmarks classifiers; no external real-world criterion validation.
    \item \textbf{D7 (Confounding): \textsc{No}.} No confound-isolation or causal identification methods.
\end{itemize}

\paragraph{Paper 11: \citeauthor{feng-etal-2023-pretraining} (ACL \citeyear{feng-etal-2023-pretraining}) --- Political Bias from Partisan Pretraining (PCT axes).}
\begin{itemize}
    \item \textbf{D1 (Construct): \textsc{Partial}.} Operationalizes leaning via PCT outputs; construct remains tied to instrument choice rather than independent definition.
    \item \textbf{D2 (Face): \textsc{Partial}.} Qualitative plausibility checks via examples; no formal expert content-validity study.
    \item \textbf{D3 (Reliability): \textsc{Yes}.} Reports chance-corrected agreement for a stance detector and includes prompt-robustness analyses.
    \item \textbf{D4 (Convergent): \textsc{Partial}.} Compares against media-bias ratings of pretraining sources (external reference point).
    \item \textbf{D5 (Discriminant): \textsc{No}.} No explicit tests separating ideology from superficial keyword/prompt effects.
    \item \textbf{D6 (Predictive): \textsc{Partial}.} Links pretraining bias to downstream fairness shifts on benchmark tasks; still not an external real-world criterion.
    \item \textbf{D7 (Confounding): \textsc{Partial}.} Uses controlled model-building (e.g., comparable corpora sizes) to isolate the pretraining factor; not causal identification.
\end{itemize}

\paragraph{Paper 12: \citeauthor{ch-wang-etal-2023-sociocultural} (EMNLP \citeyear{ch-wang-etal-2023-sociocultural}) --- Social Norms (Cross-cultural).}
\begin{itemize}
    \item \textbf{D1 (Construct): \textsc{Yes}.} Defines norms explicitly and situates them in cross-cultural/descriptive-norm framing.
    \item \textbf{D2 (Face): \textsc{Yes}.} Expert verification/editing plus concrete examples.
    \item \textbf{D3 (Reliability): \textsc{Partial}.} Some chance-corrected agreement is reported, but not as a core-label inter-annotator reliability statistic.
    \item \textbf{D4 (Convergent): \textsc{Partial}.} Aligns with established cross-cultural theory patterns; no independent gold-standard norm instrument correlation.
    \item \textbf{D5 (Discriminant): \textsc{No}.} No explicit empirical separation from nuisance factors.
    \item \textbf{D6 (Predictive): \textsc{No}.} No external outcome prediction beyond the dataset/task.
    \item \textbf{D7 (Confounding): \textsc{No}.} No causal/confound-isolation methods.
\end{itemize}

\paragraph{Paper 13: \citeauthor{bang-etal-2024-measuring} (ACL \citeyear{bang-etal-2024-measuring}) --- Political Bias as Stance + Framing/Style (LLM generations).}
\begin{itemize}
    \item \textbf{D1 (Construct): \textsc{Partial}.} Bias defined operationally as stance plus framing/style rather than theory-grounded construct definition.
    \item \textbf{D2 (Face): \textsc{Partial}.} Qualitative examples/sanity checks; no formal content-validity study.
    \item \textbf{D3 (Reliability): \textsc{Partial}.} Uses repeated generations and significance testing, but does not report chance-corrected reliability of the measurement procedure.
    \item \textbf{D4 (Convergent): \textsc{Partial}.} Compares against prompted anchor distributions rather than independent external instrument.
    \item \textbf{D5 (Discriminant): \textsc{Partial}.} Separates content from lexical polarity but lacks formal nuisance-confound tests.
    \item \textbf{D6 (Predictive): \textsc{No}.} Descriptive audit; no external criterion prediction.
    \item \textbf{D7 (Confounding): \textsc{Partial}.} Standardizes prompts/topics and uses anchor distributions; not causal identification.
\end{itemize}

\paragraph{Paper 14: \citeauthor{harel-canada-etal-2024-measuring} (EMNLP \citeyear{harel-canada-etal-2024-measuring}) --- Psychological Depth Scale (Stories).}
\begin{itemize}
    \item \textbf{D1 (Construct): \textsc{Yes}.} Construct grounded in literary/reader-response theory and defined independent of dataset.
    \item \textbf{D2 (Face): \textsc{Partial}.} Guidelines/training/calibration, but no formal expert content-validity audit.
    \item \textbf{D3 (Reliability): \textsc{Yes}.} Reports chance-corrected inter-rater reliability for human ratings (e.g., Krippendorff's $\alpha$).
    \item \textbf{D4 (Convergent): \textsc{Partial}.} LLM-judge scores correlate with human judgments, but not with an independent external gold-standard instrument.
    \item \textbf{D5 (Discriminant): \textsc{Partial}.} Argues beyond surface style; lacks formal nuisance-confound tests.
    \item \textbf{D6 (Predictive): \textsc{No}.} Used as evaluation rubric; no external real-world outcome prediction.
    \item \textbf{D7 (Confounding): \textsc{Partial}.} Uses design controls; no causal identification.
\end{itemize}

\paragraph{Paper 15: \citeauthor{azizov-etal-2024-safari} (EMNLP Findings \citeyear{azizov-etal-2024-safari}) --- Political Bias and Factuality (Cross-lingual; xMP).}
\begin{itemize}
    \item \textbf{D1 (Construct): \textsc{Partial}.} Targets operationalized via existing rating schemes rather than theory-defined constructs.
    \item \textbf{D2 (Face): \textsc{Partial}.} Relies on expert-provided outlet-level ratings and supplemental labels; no formal content-validity study for the resulting instrument.
    \item \textbf{D3 (Reliability): \textsc{Partial}.} Reports cross-check alignment between article samples and outlet labels; informative but not full stability analysis.
    \item \textbf{D4 (Convergent): \textsc{No}.} No clear external convergent validation reported.
    \item \textbf{D5 (Discriminant): \textsc{No}.} No explicit tests against nuisance confounds.
    \item \textbf{D6 (Predictive): \textsc{No}.} Benchmarks models to predict labels; no external outcome criterion.
    \item \textbf{D7 (Confounding): \textsc{No}.} No causal/confound-isolation methods.
\end{itemize}

\paragraph{Paper 16: \citeauthor{faulborn-etal-2025-little} (ACL \citeyear{faulborn-etal-2025-little}) --- Political Values Measurement (WVS/EVS items).}
\begin{itemize}
    \item \textbf{D1 (Construct): \textsc{Yes}.} Builds measurement around validated survey instruments and argues against ad-hoc quizzes.
    \item \textbf{D2 (Face): \textsc{Yes}.} Inherits content validity from decades of survey-item vetting.
    \item \textbf{D3 (Reliability): \textsc{Yes}.} Tests prompt sensitivity and demonstrates instability; reported as a stability diagnostic.
    \item \textbf{D4 (Convergent): \textsc{Partial}.} Motivates survey validity and population alignment, but does not provide a clean correlation-style convergent test as coded.
    \item \textbf{D5 (Discriminant): \textsc{Partial}.} Diagnoses wording/prefix contamination (nuisance sensitivity), but does not report a classic discriminant test against pre-specified nuisance factors.
    \item \textbf{D6 (Predictive): \textsc{No}.} Focus is measurement validity rather than external outcome prediction using the derived score.
    \item \textbf{D7 (Confounding): \textsc{Partial}.} Uses design-based mitigation; not causal identification.
\end{itemize}

\paragraph{Paper 17: \citeauthor{azzopardi-moshfeghi-2025-pow} (EMNLP Findings \citeyear{azzopardi-moshfeghi-2025-pow}) --- Political Overton Window (PRISM audit).}
\begin{itemize}
    \item \textbf{D1 (Construct): \textsc{Yes}.} Operationalizes Overton Window by mapping espouse/neutral/refusal across a spectrum.
    \item \textbf{D2 (Face): \textsc{Partial}.} Visualizations provide plausibility, but no formal expert content-validity study.
    \item \textbf{D3 (Reliability): \textsc{Yes}.} Reports chance-corrected agreement for the assessment procedure.
    \item \textbf{D4 (Convergent): \textsc{Partial}.} Interprets boundaries relative to expected alignment behavior; no independent ideology instrument correlation.
    \item \textbf{D5 (Discriminant): \textsc{Partial}.} Distinguishes position from acceptability conceptually; limited empirical nuisance-separation evidence.
    \item \textbf{D6 (Predictive): \textsc{No}.} Descriptive audit; no external criterion prediction.
    \item \textbf{D7 (Confounding): \textsc{Partial}.} Persona probing and demographic/prompt heuristics; no causal identification methods.
\end{itemize}

\end{document}